\documentclass{article}

\usepackage{arxiv}

\usepackage[utf8]{inputenc} 
\usepackage[T1]{fontenc}    
\usepackage{hyperref}       
\usepackage{url}            
\usepackage{booktabs}       
\usepackage{amsfonts}       
\usepackage{nicefrac}       
\usepackage{microtype}      
\usepackage{lipsum}
\usepackage{graphicx}
\usepackage{multirow} 
\usepackage{subcaption}
\usepackage{array}    
\usepackage[table]{xcolor}
\usepackage{makecell}
\usepackage{amsmath}

\graphicspath{ {./images/} }

\title{IPAD-CLIP: Teaching CLIP to Detect Image Local Perceptual Artifacts}

\author{
Juan Wang \\
  Institute of Automation\\
  Chinese Academy of Sciences\\
  Beijing,  China\\
  \texttt{jun\_wang@ia.ac.cn} \\
   \And
Xinyu Sun\\
School of Computer Science\\ 
Beijing Jiaotong University\\
Beijing, China\\
\texttt{sunxinyu@bjtu.edu.cn} \\
 \And
 Ke Zhang \\
 Department of Electronic Engineering\\ 
 Tsinghua University\\
 Beijing,  China\\
\texttt{zhangkedemon@163.com} \\
  \And
Jin Wang\\
Minzu University of China\\
Beijing,  China\\
\texttt{wangjin@muc.edu.cn}
\And
Bing Li \\
Institute of Automation\\
Chinese Academy of Sciences\\
Beijing,  China\\
\texttt{bli@nlpr.ia.ac.cn} \\
 \And
Weiming Hu\\
Institute of Automation\\
Chinese Academy of Sciences\\
Beijing,  China\\
\texttt{wmhu@nlpr.ia.ac.cn}\\
 \And
 Liang Wang\\
 OPPO Co., Ltd.‌\\
 \texttt{wangliang@oppo.com}\\
}

\begin{document}
\maketitle
\begin{abstract}
Current image quality assessment methods are heavily biased towards global distortions (e.g., noise, blur), neglecting local perceptual artifacts such as ghosting, lens flare, and moiré effects. Although significant progress has been made in artifact removal, the fundamental problem of automatic artifact detection remains largely unexplored. In this paper, we formalize the Image Perceptual Artifact Detection (IPAD) task to address this gap. We contribute a benchmark dataset comprising 3,520 artifact images, including 520 real-captured and 3,000 synthetic samples, each paired with pixel-level masks across three representative artifact categories. The core challenge of IPAD lies in the localized, subtle, and semantically weak nature of these artifacts, which makes them prone to missed detection. To overcome this, we introduce IPAD-CLIP, a novel framework built upon CLIP that enhances artifact discrimination in both textual and visual spaces while preserving generalization capabilities. Our key insight is that local artifacts often exhibit strong correlations with specific semantic contexts. Accordingly, we learn artifact-aware text embeddings to explicitly model the object-artifact relationships, resulting in enhanced representations that clear differentiate between clean and artifact prompts. These text embeddings are then used as anchors to shift the visual encoder’s attention from high-level semantics to subtle, low-level artifacts. Extensive experiments demonstrate that IPAD-CLIP offers a resource-efficient adaptation of CLIP for detection, significantly outperforming advanced image anomaly detection and manipulation detection methods on our benchmark. To the best of our knowledge, this is the first study addressing multi-class local perceptual artifact detection in terms of both dataset and model.
\end{abstract}


\section{Introduction}

Image quality assessment (IQA) is a fundamental task in computer vision that aims to evaluate image quality in alignment with human perception. Its importance has grown rapidly, as IQA models increasingly serve as critical reward signals to guide imaging and image enhancement algorithms. Current IQA methods predominantly focus on global distortions, such as noise, blur, and compression artifacts. However, they largely overlook local artifacts, including ghosting, lens flare, and moiré, which typically appear at object boundaries, around bright light sources, and on screens, respectively. These artifacts can be distracting, reduce detail, and occlude image content, degrading both human perceptual experience and machine vision performance. 
Although some works have focused on removing these artifacts \cite{flare7k, wu2021train, doingMore, HDR_recon} under the assumption that they are already present, there is a notable lack of frameworks capable of automatically detecting them. This limitation forces manual judgment to determine the presence of artifacts or compels the application of removal algorithms to all images, leading to inefficient workflows and potential over-processing. Therefore, accurately detecting these artifacts is crucial for comprehensively diagnosing image quality and for optimizing imaging and enhancement algorithms to prevent their occurrence.

\begin{figure}[tbp]
  \centering
  \includegraphics[width=0.7\linewidth]{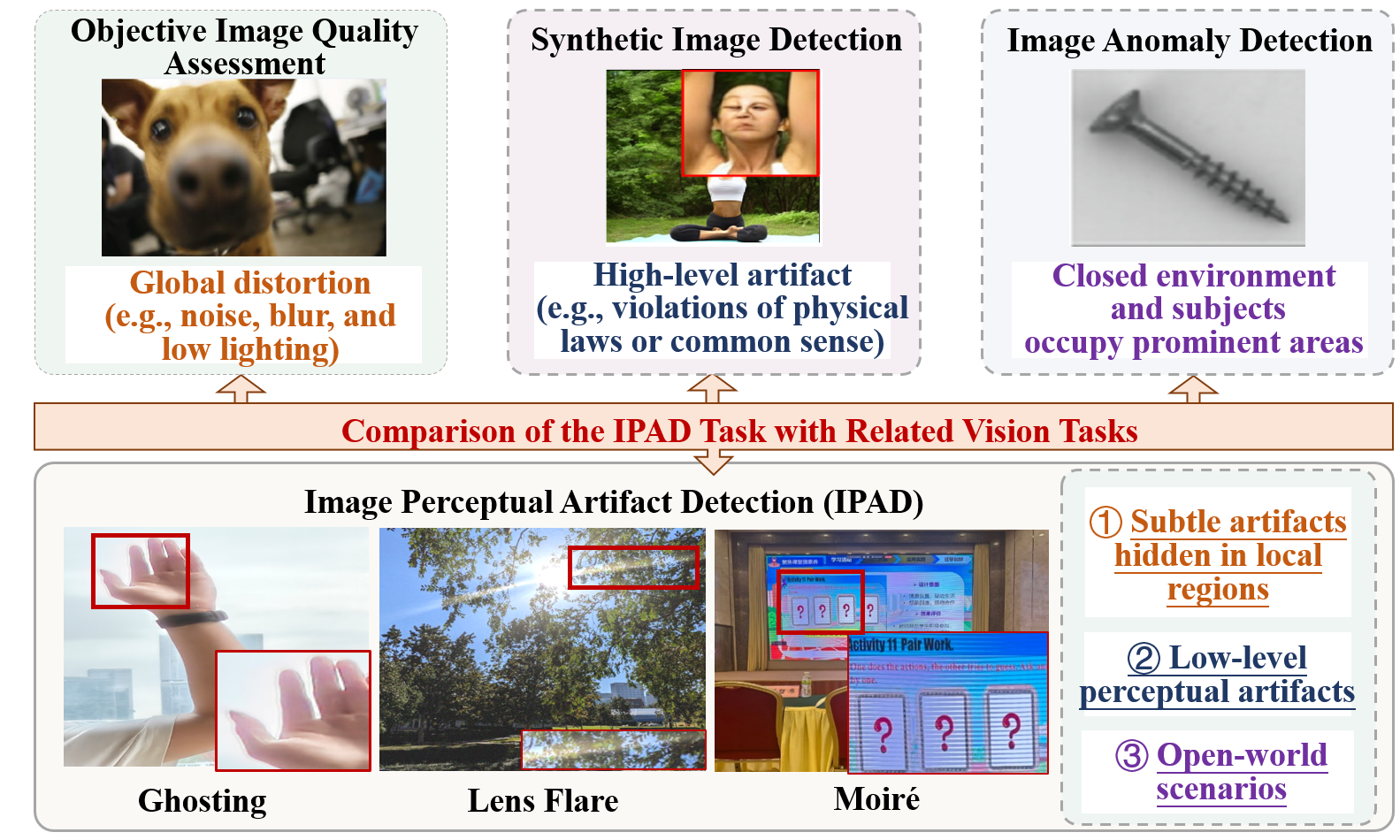}
  \caption{Comparison of the Image Perceptual Artifact Detection (IPAD) task with related vision tasks. IPAD focuses on the precise classification and localization of low-level perceptual artifacts within diverse, open-world natural scenes.
  }
 \label{fig:distinction}
  \vspace{-0.3cm}
\end{figure}

To pioneer research in this direction, we propose a novel benchmark task, Image Perceptual Artifact Detection (IPAD), which aims to accurately classify artifact types and localize artifact regions. While related to existing tasks, this task possesses distinct characteristics, as shown in Figure \ref{fig:distinction}. First, unlike objective IQA \cite{wjhierarchical, promptiqa, chen2022teacher}, which estimates global distortion levels and predicts a scalar score, this task requires the precise classification and localization of local artifacts. Second, in contrast to synthetic image detection \cite{psccnet, mvss, lgrad, freqnet}, which typically identifies high-level semantic artifacts (e.g., violations of physical laws or common sense), this task focuses on low-level perceptual artifacts. Third, unlike anomaly detection \cite{mambaad, ViTAD, uniAD, dinomaly}, which is often restricted to controlled, closed-set domains like industrial or medical imaging, this task operates in diverse, open-world natural scenes. These unique characteristics render existing methods ill-suited for this task, necessitating the development of specialized frameworks. 

The challenges of the IPAD task are twofold.
First, there is a lack of training data. Although prior research has explored artifact removal and created related datasets \cite{flare7k, wu2021train, doingMore, HDR_recon}, these datasets either lack corresponding masks to annotate artifact locations or rely on synthetic artifacts that are global rather than localized. To address this gap, we construct the first multi-class image perceptual artifact dataset, comprising three representative artifact types: ghosting, lens flare, and moiré. Our dataset includes 520 real-captured artifact images and 3,000 synthetic artifact images, each accompanied by a corresponding mask that accurately annotates the artifact regions.

Second, these artifacts are localized, subtle, and exhibit diverse appearances. Adopting a vision-only supervised learning paradigm risks learning spurious correlations from limited visual patterns without understanding the concept of artifacts, making them prone to being missed detection. To overcome this, we introduce IPAD-CLIP, a novel framework built upon CLIP \cite{clip} that enhances artifact discrimination in both textual and visual spaces while preserving generalization capabilities. Our key insight is that these local artifacts are closely related to specific semantics. For instance, ghosting often appears at the edges of subjects, lens flare frequently occurs around light sources, and moiré commonly arises on screens. Based on this observation, we propose to leverage CLIP's pre-trained multi-modal semantic priors to ground artifact detection. Specifically, we employ learned text embeddings to implicitly model object–artifact relationships, creating "anchors" for artifact-aware semantics within the text space. These enhanced text embeddings then guide CLIP's vision encoder to shift its attention from semantic regions to artifact regions. This new paradigm constrains the visual feature space to be interpretable and discriminative. 

Moreover, to preserve CLIP's pre-trained knowledge for robust generalization across diverse scenes, we incorporate lightweight adapters. This design enables controlled adaptation of CLIP, enhancing its capability to handle the fine-grained IPAD task without sacrificing generalization. Additionally, we utilize both image-level and pixel-level loss functions, enabling the model to perceive artifacts globally and locally. Experimental results on our benchmark dataset demonstrate a significant advantage of our method over advanced image manipulation detection and anomaly detection approaches. To the best of our knowledge, this is the first dataset and framework for holistic detection of multi-class local artifacts.

In summary, this paper makes the following main contributions:

(1) We introduce a novel task: Image Perceptual Artifact Detection (IPAD). To facilitate research in this direction, we construct the first multi-class image perceptual artifact dataset, which enables training of artifact detection models and provides a comprehensive evaluation benchmark.

(2) We propose IPAD-CLIP, a framework that enhances CLIP's capability for local artifact detection. We design learnable text anchors that encode artifact-aware semantics to align with multi-granularity image features, enabling fine-grained artifact perception. Additionally, we incorporate lightweight adapters that boost detection performance while preserving CLIP's generalization.

(3) Comprehensive experiments demonstrate that IPAD-CLIP achieves consistently superior performance in both artifact classification and segmentation, outperforming state-of-the-art image manipulation detection and anomaly detection methods.

\section{Related Work}

In this section, we introduce three related vision tasks to image perception artifact detection.

\subsection{Objective Image Quality Assessment}
Objective image quality assessment (IQA) primarily focuses on evaluating global image distortions, such as noise, blur, and artifacts, typically requiring models to predict a Mean Opinion Score (MOS). Initially, methods relied on hand-crafted natural image statistics \cite{iqa1}. Subsequently, deep learning-based approaches \cite{chen2022teacher} superseded these manual features by learning quality priors from extensive datasets. Further improvements have been achieved by incorporating curriculum learning \cite{wjhierarchical}, prompt learning \cite{promptiqa}, and multi-task learning \cite{zhang2023blind}. However, a single scalar score offers limited information. Consequently, recent research has introduced Vision-Language Models (VLMs) to leverage their cross-modal priors. These models enable more flexible, interactive, and explainable IQA frameworks, supporting fine-grained attribute assessment \cite{clip-iqa}, localized region-based evaluation \cite{chen2024q, seagull}, multi-image comparison \cite{zhu2024adaptive}, and natural language quality descriptions \cite{qbench, qinstruct}.

\subsection{Synthetic Image Detection}
From GANs \cite{gans} to diffusion \cite{DDPM}, image generation technologies have evolved rapidly, producing increasingly diverse and detailed synthetic content. Nevertheless, synthetic images often contain implausible details, such as fake textures or violations of physical laws, that render them less realistic. Synthetic image detection task aims to discriminate between synthetic and real images. Early approaches treated this as a binary classification task, leveraging features from the spatial or frequency domains \cite{freqnet, deepfake1}. However, these methods frequently struggle with generalization across diverse generators and robustness against various perturbations; more importantly, they lack interpretability. Recently, many works have extended the initial binary classification paradigm to the more complex task of artifact localization. For instance, some studies \cite{truFor, hifinet}  have utilized gradients or attention maps to reveal potential anomalous regions, while others \cite{legion} have focused on constructing datasets annotated with detailed artifact segmentation masks.

\subsection{Image Anomaly Detection}
Image anomaly detection techniques are widely applied in medical and industrial fields, aiming to locate lesions in medical images or anomalies on industrial products. Early research primarily focused on the one-class–one-model setting, which faces increasing limitations in terms of deployment costs. Consequently, research has shifted toward unified multi-class anomaly detection. Representative works such as UniAD \cite{uniAD} first systematically explored single-model processing for multi-class anomaly detection; MambaAD \cite{mambaad} introduced state space models to enhance global and local pattern modeling capabilities; and Dinomaly \cite{dinomaly} alleviated the identity mapping problem under the unified setting through a more concise reconstruction framework. Few-shot anomaly detection has also gained widespread attention in recent years, such as AA-CLIP \cite{aaclip} and AnomalyCLIP \cite{anomalyclip}, which enables unified anomaly detection across different domains (e.g., industrial and medical) by requiring only a few normal reference samples during the inference.

\section{Dataset Construction}

\begin{figure}[tbp]
  \centering
  \includegraphics[width=0.7\linewidth]{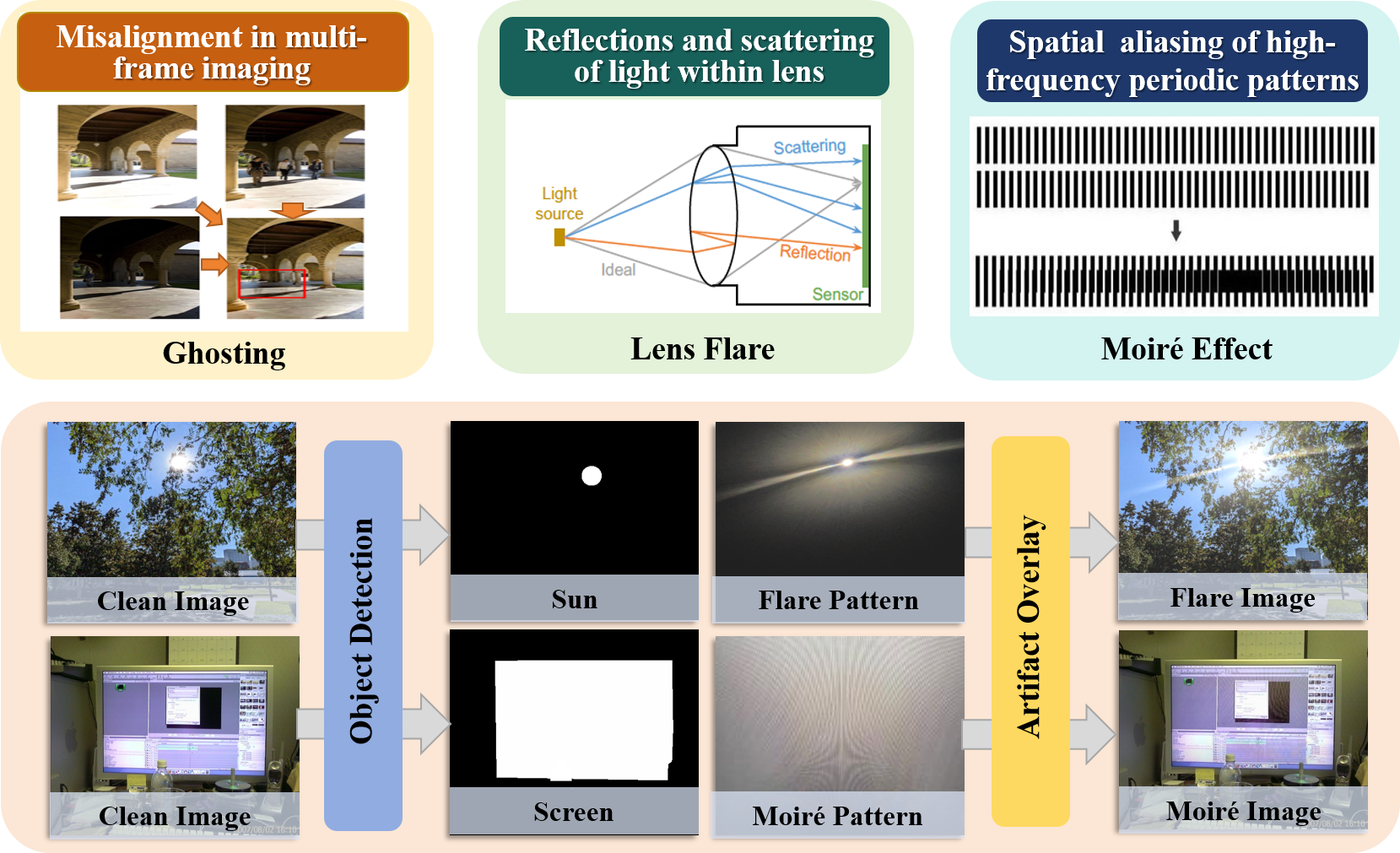}
  \caption{Physical origins of three artifacts (top), with augmentation via multi-frame fusion for ghosting and pattern-based synthesis for flare and moiré (bottom).}
 \label{fig:synthetic}
  \vspace{-0.3cm}
\end{figure}

\subsection{Physical Origins of Artifacts}
Our dataset comprises three representative types of perceptual artifacts: ghosting, lens flare, and moiré effects. Each artifact arises from distinct physical mechanisms (as shown in Figure \ref{fig:synthetic}) and presents unique visual characteristics, posing diverse challenges for detection methods.

\textbf{Ghosting:} Ghosting artifacts frequently occur when relative motion exists between the camera and the subject during exposure. This motion causes misalignment among captured frames, resulting in semi-transparent, duplicate-like artifacts in the final composite image. Such artifacts are commonly observed in multi-frame imaging techniques, including High Dynamic Range (HDR) merging, noise reduction, and super-resolution, where multiple frames are aligned and fused to produce a single output. Imperfect alignment in these processes often manifests as ghosting, particularly in scenes containing moving subjects.

\textbf{Lens Flare:} Lens flare is a prominent visual artifact that arises when photographs are captured in the presence of a strong light source. It results from unintended internal reflections and scattering within the camera lens system, often producing characteristic streaks, circles, or haze across the image. These artifacts can obscure scene details and degrade both visual quality and the reliability of downstream computer vision tasks.

\textbf{Moiré Effects:} Moiré effects stem from spatial aliasing caused by interference between the high-frequency periodic patterns of a device screen, such as its pixel grid or color filter array—and the sampling lattice of the camera sensor. This phenomenon occurs when the spatial frequency of the scene exceeds the sensor's Nyquist limit, giving rise to visible, low-frequency beat patterns that were not present in the original scene. Moiré artifacts are particularly prevalent in images captured from digital displays, posing significant challenges for applications involving screen content.

\subsection{Data Collection}
We first collect 520 real-captured artifact images, comprising 150 ghosting, 192 lens flare, and 178 moiré samples. To augment the dataset, we generate  additional  artifact images. 
For ghosting, we adopt the protocol described in \cite{Kalan} to acquire image sequences, which are then processed using various multi-frame fusion models, such HDR algorithms \cite{hdr1, hdr2}. The resulting ghosting regions are then carefully annotated by four volunteers.
For flare and moiré artifacts, we leverage diverse artifact patterns from prior work \cite{flare7k, wu2021train, doingMore, HDR_recon} to generate synthetic artifact images. As shown in Figure \ref{fig:synthetic}, given a clean image $I_{\text{gt}} \in \mathbb{R}^{H \times W \times 3}$, we first detect the sun and the screen in the image, producing binary masks. We then synthesize the artifact image by overlaying lens flare patterns onto the detected sun region and overlaying moiré patterns onto the detected screen region.
Formally, let  $N \in \mathbb{R}^{H \times W \times 3}$ 
denote the artifact pattern, and $M \in \mathbb{R}^{H \times W}$ be a binary mask indicating the region where the artifact is applied. $M(i,j)=1$  indicates the presence of an artifact and $M(i,j)=0 $ denotes a clean region. The  synthetic artifact image $I$ is generated as:
\begin{equation}
I = I_{\text{gt}} \odot (1-M) + [(1 - \phi) I_{\text{gt}}+ \phi N] \odot  M,
\end{equation}
where $\odot$ denotes the Hadamard product and $0 < \phi <1$ controls the blending intensity. For lens flare and moiré effects, we use separate masks $M_{\text{sun}}$ and $M_{\text{screen}}$ with their respective patterns and different blending factors $\phi_f$ and $\phi_m$.

In addition, our dataset contains 1,000 clean images without any artifacts. This is important because artifacts frequently appear in specific semantic contexts; including artifact-free images encourages the model to learn robust discrimination between artifact and clean samples rather than relying on spurious semantic correlations.

\begin{figure}[bp]
  \centering
  \includegraphics[width=0.7\linewidth]{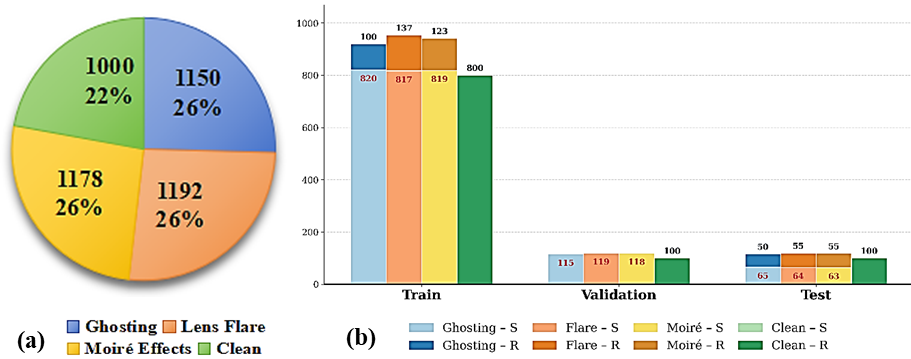}
  \caption{ (a) Image counts for clean images and each artifact type. (b) Image counts per subset, where "S" indicates synthetic images and "R" indicates real-captured images.}
 \label{fig:dataset}
  \vspace{-0.3cm}
\end{figure}

\begin{figure*}[h]
  \centering
  \includegraphics[width=0.9\linewidth]{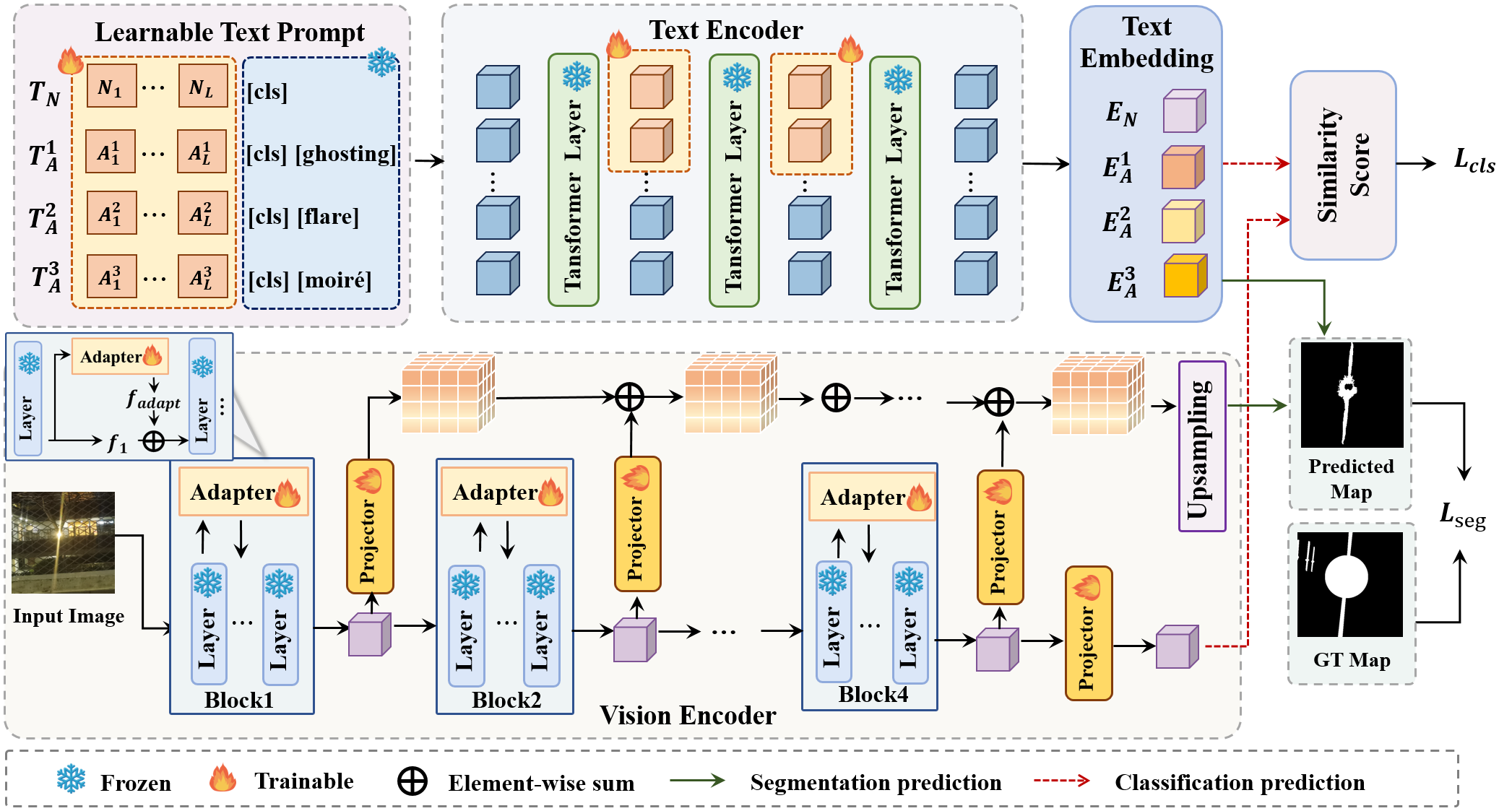}
  \caption{Overview of the IPAD-CLIP framework. We learn artifact-aware text embeddings to capture object-artifact relationships, which then adapt CLIP's vision encoder to shift attention from object semantics to fine-grained artifacts.}
 \label{fig:framework}
  \vspace{-0.3cm}
\end{figure*}

Table \ref{fig:dataset} provides a detailed breakdown of our dataset, including the number of images for clean and each artifact type, the number for training, validation and testing subsets, and the number of real-captured and synthetic data.

\section{Proposed Method}

\subsection{Overview}

We formally define the task of Image Perceptual Artifact Detection (IPAD). Given an input image $I \in \mathbb{R}^{W \times H \times 3}$, the goal is twofold: (1) predict the artifact class $k \in \{0, 1, \dots, K\}$, where $k = 0$ indicates a clean image and $k \in \{1, \dots, K\}$ denotes one of $K$ artifact types; and (2) generate a binary mask $M \in \{0, 1\}^{W \times H}$ that localizes the artifact region. For clean images ($k = 0$), the mask is defined as an all-zero matrix.

The most significant challenge of the IPAD task lies in that these artifacts are localized, subtle, and exhibit significant appearance variations, making them prone to being missed during detection.
Motivated by the observation that there exists a  correlation between objects and artifacts, we implicitly model this relationship using text priors to facilitate artifact recognition. To this end, we propose IPAD-CLIP, a framework designed to enhance CLIP's ability to discriminate between artifact and clean images while preserving its pre-trained knowledge for generalization to open-world scenes.
An overview of IPAD-CLIP is presented in Figure \ref{fig:framework}. 
Our framework employs a three-stage training strategy to progressively learn artifact-aware representations. Stage I adapts the CLIP vision encoder to establish a task-specific visual foundation. Stage II learns artifact-aware text embeddings to enhance discrimination between clean and artifact anchors in the text space. Stage III aligns image features with these text embeddings, shifting the vision encoder's attention from object semantics to fine-grained artifacts.

\subsection{Learning Artifact-aware Text Embeddings}

Common CLIP text prompt templates, such as "a photo of a [cls]", are primarily designed to capture object-level semantics and thus struggle to produce textual embeddings that effectively discriminate between clean and artifact regions. A naive solution is using templates like "a photo of a [cls] with [artifact]". However, this strategy is suboptimal, as CLIP has limited prior knowledge of artifact-specific terms such as "ghosting," "flare," and "moiré."

To address this, we introduce learnable text prompts that are optimized to encode both artifact information and contextual semantic cues. These prompts consist of learnable word embeddings that adapt to the target domain, yielding highly discriminative representations between clean and artifact regions. Formally, for the clean prompt and the $k$-th artifact type, we define:
\begin{align}
    T_N &= [N_1][N_2]\dots[N_L][\text{cls}], \\
    T^k_A &= [A^k_1][A^k_2]\dots[A^k_L][\text{cls}][\text{artifact}^k],
\end{align}
where [cls] denotes the object-level semantic description (e.g., "a photo of [object]"), $[\text{artifact}^k]$ indicates the $k$-th artifact type (e.g., "ghosting"), and $[N_i]$ and $[A^k_i]$ are learnable word embeddings of length $L$ for the clean and artifact prompts, respectively. The text encoder then produces corresponding embeddings $E_N$ and $\{E_A^k\}_{k=1}^{K}$ for subsequent multi-modal alignment.

To facilitate learning a more discriminative textual space adapted to the IPAD task, we introduce learnable token embeddings into the text encoder \cite{anomalyclip, maple}. Specifically, we replace the prefix of the original token embeddings at each layer with learnable tokens, enabling layer-wise adaptation.
Formally, let $d_h = [d_h^1, d_h^2, \dots, d_h^V]$ denote the token embeddings at the $h$-th layer of the text encoder, where $V$ is the sequence length. We introduce $J$ learnable tokens $g_h = [g_h^1, g_h^2, \dots, g_h^J]$ with $J < V$, and replace the first $J$ tokens of $d_h$ to obtain the adapted input:
\begin{equation}
    d'_h = \{g_h^1, g_h^2, \dots, g_h^J, d_h^{J+1}, \dots, d_h^V\}.
\end{equation}
This adapted sequence $d'_h$ is then processed by the $h$-th transformer layer, producing output:
\begin{equation}
    d_{h+1} = \{o_{h+1}^1, o_{h+1}^2, \dots, o_{h+1}^J, d_{h+1}^{J+1}, \dots, d_{h+1}^V\},
\end{equation}
where $\{o_{h+1}^j\}_{j=1}^{J}$ are the transformed representations of the learnable tokens. To maintain layer-wise adaptation, we discard these transformed tokens and reinitialize a new set of learnable embeddings $g_{h+1} = [g_{h+1}^1, \dots, g_{h+1}^J]$ for the next layer. The input to layer $h+1$ is then constructed as:
\begin{equation}
    d'_{h+1} = [g^1_{h+1},...,g^J_{h+1}, d_{h+1}^{J+1},...,d_{h+1}^V].
\end{equation}
This process is repeated across all designated layers, allowing the learnable prompts to progressively refine textual representations for artifact discrimination.

\subsection{Learning Multi-granularity Vision Features}

To preserve CLIP's pre-trained knowledge while enabling targeted adaptation for artifact detection, we introduce lightweight adapters into the first $L$ layers of the vision encoder. For the $i$-th transformer layer ($i \leq L$), the output feature $F^i \in \mathbb{R}^{N \times d}$ is fed into a trainable adapter, producing an adapted feature:
\begin{equation}
    F^i_{adapt} = \text{Norm}\big(\text{Act}(W^i F^i)\big),
    \label{eq:adapter}
\end{equation}
where $W^i \in \mathbb{R}^{q \times q}$ is the trainable linear weight of the $i$-th adapter, $\text{Act}(\cdot)$ is an activation function, and $\text{Norm}(\cdot)$ denotes layer normalization. To preserve the original CLIP's generalization while incorporating artifact-specific cues, we fuse the original and adapted features via a weighted residual connection:
\begin{equation}
    \tilde F^i = \beta \, F^i_{adapt} + (1 - \beta) \, F^i,
    \label{eq:fusion}
\end{equation}
where $\beta \in [0,1]$ controls the trade-off between preserving pre-trained knowledge and adapting to artifact patterns. 
In addition to layer-wise adaptation, we leverage multi-granularity features to enhance artifact localization, as subtle artifacts often manifest at varying spatial scales. Specifically, we extract intermediate features $\{\tilde F^i\}_{i \in \mathcal{I}}$ from a set of selected layers $\mathcal{I}$, capturing representations at different granularities. Each feature $\tilde F^i$ is passed through a trainable projector $\text{Proj}_i(\cdot)$ to align its channel dimension with the text embedding space, yielding $F^i_{MG} = \text{Proj}_i(\tilde F^i)$. The final multi-granularity representation is obtained by aggregating these projected features:
\begin{equation}
    F_{MG} = \sum_{i \in \mathcal{I}} F_{MG}^i= \sum_{i \in \mathcal{I}} \text{Proj}_i(\tilde F^i).
    \label{eq:patch_aggregation}
\end{equation}
This aggregated representation captures complementary multi-scale information, enabling robust detection of artifacts with varying spatial extents.

\subsection{Three-stage Training}
\textbf{Stage I: Initial Vision Adaptation.} 
We freeze the pre-trained CLIP vision encoder and train only the adapters and projectors, without leveraging text embeddings. Given an input image $I$, the vision encoder with trainable adapters extracts both the global latent code $F_{image}$ and the multi-granularity features $F_{MG}$. A classification head and a segmentation head, implemented as fully-connected layers, are then applied to produce the predicted class probability $\hat{y}_c$ and segmentation mask $\hat{S}$.
The classification loss $\mathcal{L}_{cls}$ and segmentation loss $\mathcal{L}_{seg}$ are computed to optimize the trainable parameters. Specifically, the classification loss employs cross-entropy (CE):
\begin{equation}
    \mathcal{L}_{cls} = - \sum_{c=0}^{K} \mathbb{I}(y = c) \cdot \log(\hat y_c),
    \label{eq:multiclass_loss}
\end{equation}
where $y$ is the ground-truth (GT) artifact type of image $I$, $\hat{y}_c$ is the predicted probability for class $c$, and $\mathbb{I}(\cdot)$ is the indicator function.
The segmentation loss combines Dice loss and Focal loss applied to the predicted mask $\hat{S}$ and the GT mask $S$. The total loss is:
\begin{equation}
   \begin{aligned}
        \mathcal{L}_{total} & = \lambda \mathcal{L}_{cls} + \mathcal{L}_{seg}\\
         & = \lambda \cdot \text{CE}(\hat{y}, y) + \text{Dice}(\hat{S}, S) + \text{Focal}(\hat{S}, S),
 \end{aligned}
    \label{eq:losses}
\end{equation}
where $\lambda$ is a hyperparameter balancing the two losses.

\textbf{Stage II: Text Embeddings Learning.}
Building upon the vision features adapted in Stage I, we now train the learnable text embeddings while freezing all other parameters. Given an input image $I$, we extract the global feature $F_{image}$ and multi-granularity features $F_{MG}$ using the vision encoder adapted in Stage I. We then compute cosine similarity between these visual features and the text embeddings $[E_N, E^1_A, \dots, E^K_A]$. For notational simplicity, we index these embeddings as $[E^k]_{k=0}^K$, where $k=0$ represents the clean prompt and $k=\{1,\dots,K\}$ represents the $k$-th artifact type. The classification and segmentation predictions are obtained as:

\begin{equation}
\begin{aligned}
    \hat{y}_c &= \frac{\exp\left( \cos\left( F_{image}, E^c \right) \right)}{\sum_{j=0}^{K} \exp\left( \cos\left( F_{image}, E_j \right) \right)},\\
    \hat{S}_c(p) &= \frac{\exp\left( \cos\left( F_{MG}(p), E^c \right) \right)}{\sum_{j=0}^{K} \exp\left( \cos\left( F_{MG}(p), E_j \right) \right)},
    \label{eq:predict}
\end{aligned}
\end{equation}
where $p$ indexes spatial positions, $F_{MG}(p)$ is the feature vector at position $p$, $\hat{y}_c$ is the image-level probability for class $c$, and $\hat{S}_c(p)$ is the pixel-level probability that position $p$ belongs to class $c$. 

\begin{figure}[htbp]
  \centering
  \includegraphics[width=0.7\linewidth]{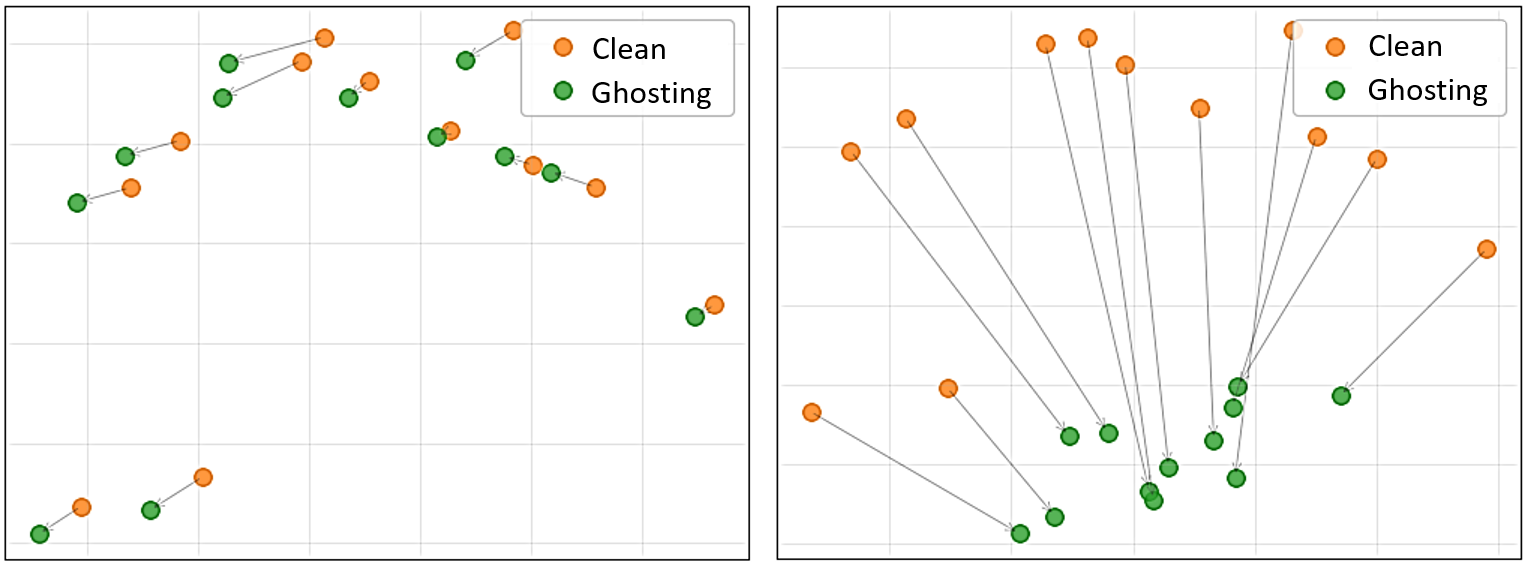}
  \caption{T-SNE visualization comparison of CLIP's text embeddings for clean and artifact text prompts before (left) and after (right) the artifact-aware text embedding learning. }
 \label{fig:clip}
  \vspace{-0.3cm}
\end{figure}

The loss function follows the same formulation as Eq.~(\ref{eq:losses}), substituting the predictions from Eq.~(\ref{eq:predict}). These two terms jointly supervise the learning process at the image and pixel levels, guiding the text embeddings to adaptively reshape toward the actual visual distribution of artifacts. Figure~\ref{fig:clip} presents a t-SNE visualization of the text embeddings for clean–artifact prompt pairs. The results show that, compared to the original CLIP's text embeddings (left), the learned text embeddings (right) achieve clearer separation between clean and artifact prompts, thereby establishing higher-quality text anchors.

\textbf{Stage III: Multi-modal Alignment for Vision Enhancement.}
Although the vision encoder was adapted in Stage I, the purely visual model still struggles to distinguish artifacts due to spurious correlations learned from limited visual patterns without understanding the underlying artifact concepts, leading to high false positive rates. In the third stage, we leverage the learned text embeddings to refine the vision features. Specifically, we freeze the text embeddings $[E^k]_{k=0}^K$ obtained from Stage II and fine-tune only the vision adapters and projectors, aligning the image features with the artifact-aware text embeddings.

The alignment is guided by the loss function in Eq.~(\ref{eq:losses}), with predictions computed via Eq.~(\ref{eq:predict}). The learned text embeddings serve as anchors, constraining the visual feature space to be both globally and locally interpretable and discriminative. This multi-modal guidance is particularly beneficial when the visual signal is weak, as it helps the model focus on artifact-relevant features while suppressing task-irrelevant variations.



\section{Experiments}

\subsection{Experiment Setups}

\textbf{Implementation Details.}
We employ OpenCLIP\footnote{https://github.com/mlfoundations/open clip} with ViT-L/14 as our backbone, with all its parameters frozen during training. The input images are resized to $518 \times 518$ with batch size of 8.
The learnable token embeddings are prepended to the inputs of the first 9 layers of the text encoder to refine the textual representation space. The adapters are added to the first 6 layers of the visual encoder. We set $\beta=0.1$, $\lambda=4$ and $K=3$. We extract the outputs of the 6th, 12th, 18th, and 24th layers of the vision encoder to construct the multi-granularity features. Adam optimizer is utilized with learning rate of $1\times10^{-3}$. The network is trained for 20 epochs. All experiments are conducted on a single NVIDIA GeForce RTX 3090 GPU.

\textbf{Evaluation Metrics.}
We adopt 7 evaluation metrics. Classification performance is measured by the Area Under the Receiver Operator Curve (AUROC), Average Precision (AP), and F1 score under optimal threshold (F1-max). Segmentation performance is measured by AUROC, AP, F1-max and the Area Under the Per-Region-Overlap (AUPRO). 

\subsection{Comparison with SOTA Methods}

\begin{table*}[ht]
\centering
\caption{Comparison with anomaly detection and image manipulation detection methods across \underline{C}lassification and \underline{S}egmentation metrics. \textit{Green: best; yellow: second best}}
\label{tab:comparison}
\renewcommand{\arraystretch}{1.2}
\begin{tabular}{l c c c c c c c c}
\toprule
\rowcolor{gray!6}
 \multicolumn{2}{c}{\textbf{Method}} & \textbf{C-AUROC} & \textbf{C-AP} & \textbf{C-F1} & \textbf{S-AUROC} & \textbf{S-AP} & \textbf{S-F1} & \textbf{S-AUPRO} \\
\cmidrule(lr){1-2}\cmidrule(lr){3-5}\cmidrule(lr){6-9}
\multirow{5}{*}{\shortstack{\textbf{Anomaly}\\ \textbf{Detection}}} 
& \textbf{MambaAD} \cite{mambaad}  & 61.41 & 59.42 & 60.42 & 79.85 & 20.28 & 23.47 & 61.14 \\
& \textbf{ViTAD} \cite{ViTAD}   & 63.77 & 60.17 & 66.92 & 84.82 & 22.55 & 25.24 & 62.49 \\
& \textbf{UniAD} \cite{uniAD}    & 66.71 & 65.36 & 68.23 & 85.02 & 27.76 & 29.80 & 69.70 \\
& \textbf{Dinomaly} \cite{dinomaly}  & 67.24 & 65.82 & 69.18 & \cellcolor{yellow!25}\textbf{89.36} & \cellcolor{yellow!25}\textbf{30.89} & \cellcolor{yellow!25}\textbf{32.04} & \cellcolor{yellow!25}\textbf{74.13} \\
\midrule
\multirow{4}{*}{\shortstack{\textbf{Manipulation}\\ \textbf{Detection}}}
& \textbf{PSCCNet} \cite{psccnet} & 65.07 & 64.67 & 66.21 & 81.25 & 23.48 & 26.42 & 63.47 \\
& \textbf{MVSS-Net} \cite{mvss} & 66.18 & 64.74 & 67.79 & 80.03 & 24.22 & 25.76 & 67.68 \\
& \textbf{LGrad}  \cite{lgrad}   & 69.97 & 67.43 & 65.23 & 87.69 & 27.78 & 27.89 & 62.81 \\
& \textbf{FreqNet}\cite{freqnet}  & \cellcolor{yellow!25}\textbf{73.21} & \cellcolor{yellow!25}\textbf{70.69} & \cellcolor{yellow!25}\textbf{72.14} & 88.74 & 29.96 & 31.13 & 72.37 \\
\midrule
\rowcolor{green!8}
\textbf{Proposed} & \textbf{IPAD-CLIP} & \textbf{82.31} & \textbf{79.20} & \textbf{77.12} & \textbf{93.10} &\textbf{41.66} & \textbf{43.14} & \textbf{82.87} \\
\bottomrule
\end{tabular}
\end{table*}

We compare our method against four image anomaly detection methods, including MambaAD \cite{mambaad}, VITAD \cite{ViTAD}, UniAD \cite{uniAD}, Dinomaly \cite{dinomaly},  and  four image manipulation detection methods, including PSCCNet \cite{psccnet}, MVSS-Net \cite{mvss}, LGrad \cite{lgrad}, FreqNet \cite{freqnet}.

\begin{figure*}[htbp]
  \centering
  \includegraphics[width=\linewidth]{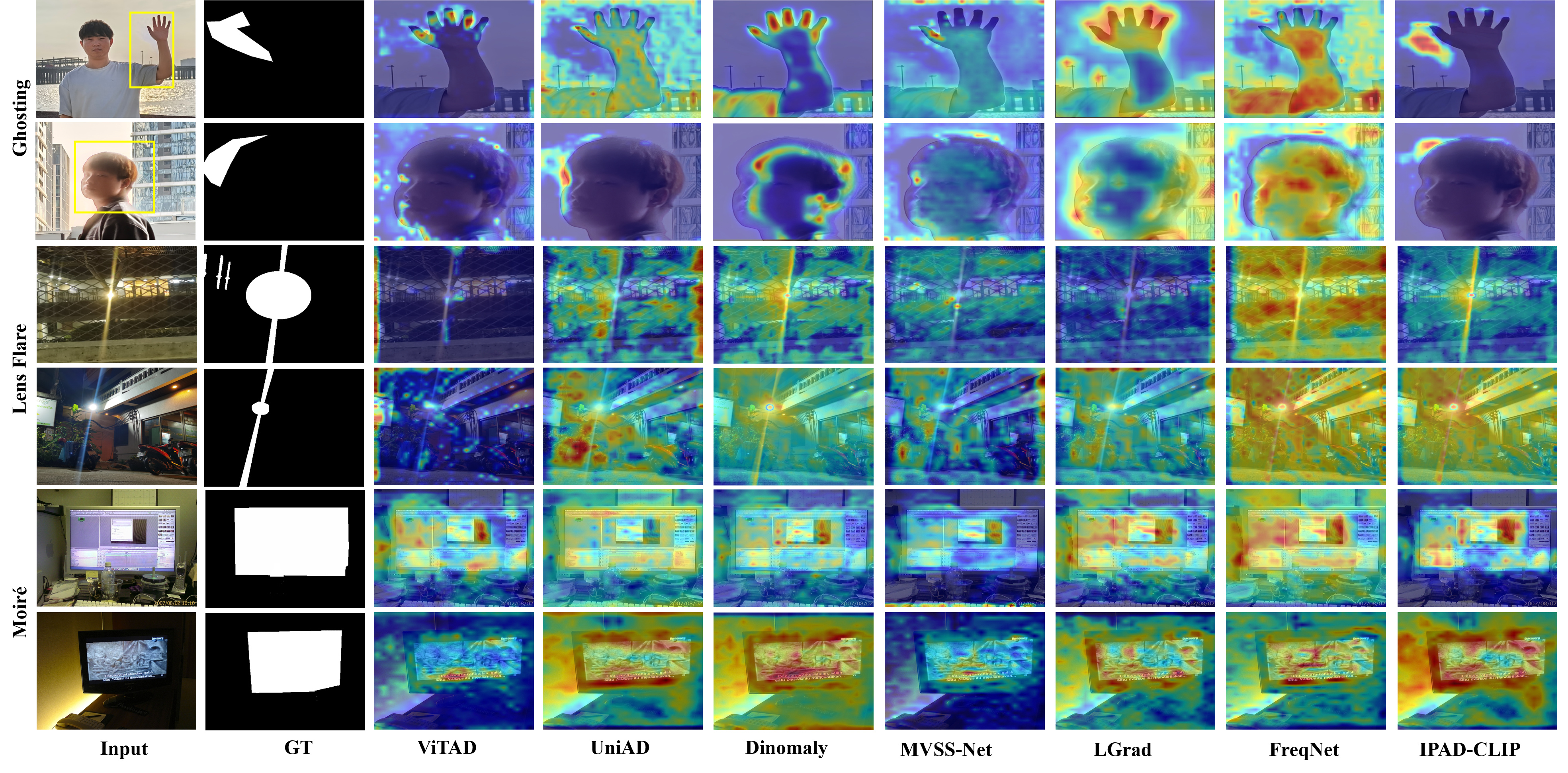}
  \caption{Visualization of artifact detection results of  ViTAD \cite{ViTAD}, UniAD \cite{uniAD}, Dinomaly \cite{dinomaly}, MVSS-Net \cite{mvss}, LGrad \cite{lgrad}, FreqNet \cite{freqnet} and our IPAD-CLIP. }
 \label{fig:subj}
  \vspace{-0.3cm}
\end{figure*}

Quantitative results are presented in Table \ref{tab:comparison}. Our IPAD-CLIP consistently outperforms all compared methods by a substantial margin on both classification and segmentation metrics, demonstrating its superior capability. Anomaly detection methods struggle to capture subtle artifacts in complex scenes, and manipulation detection methods fail to recognize artifacts that blend with scene content. In contrast, our IPAD-CLIP leverages two key priors to address these challenges: object-artifact prior and clean-artifact prior, which enhance the model's sensitivity to subtle artifact patterns.

Visualizations are demonstrated in Figure \ref{fig:subj}. As shown, most compared methods fail to attend to artifacts. For example, in the first two rows, they primarily focus on the hand and head rather than the ghosting artifacts appearing at their edges. Moreover, these methods struggle to accurately detect lens flare, likely due to its diverse appearances and ambiguous boundaries. Similarly, many methods fail to precisely recognize subtle moiré patterns on the screen. In contrast, our IPAD-CLIP demonstrates superior detection performance, effectively localizing these challenging artifacts and confirming that our framework successfully learns artifact patterns rather than being distracted by object semantics.

\subsection{Ablation Study}

In this section, we conduct ablation studies from four perspectives: text embeddings, vision encoder, training strategy, and dataset.

\begin{table}[htbp]
    \centering
    \caption{Ablation study on the components of IPAD-CLIP.}
    \label{tab:ablation_study}
    \begin{tabular}{l c c c c c c}
        \toprule
        \multirow{2}{*}{\textbf{Variant}} & \multicolumn{3}{c}{\textbf{Classification}} & \multicolumn{3}{c}{\textbf{Segmentation}} \\ 
        \cmidrule(lr){2-4} \cmidrule(lr){5-7}
        & \textbf{AUROC} & \textbf{AP} & \textbf{F1} & \textbf{AUROC} & \textbf{AP} & \textbf{F1} \\
        \midrule
        \rowcolor{gray!10} \multicolumn{7}{c}{\textbf{Text Embedding}} \\
        w/o LTE   & 71.66 & 70.37 & 72.79 & 87.76 & 30.46 & 31.82 \\
        w/o CLS   & 80.19 & 78.06 & 76.45 & 91.37 & 39.48 & 37.85 \\
        \rowcolor{gray!10} \multicolumn{7}{c}{\textbf{Vision Encoder}} \\
        w/o AD    & 67.01 & 68.44 & 69.47 & 87.28 & 29.64 & 30.79 \\
        w/o MG    & 78.84 & 74.23 & 71.69 & 88.42 & 31.69 & 34.72 \\
        \rowcolor{gray!10} \multicolumn{7}{c}{\textbf{Training Strategy}} \\
        w/o S-II-III & 70.28 & 70.96 & 71.48 & 86.59 & 29.37 & 31.01 \\
        w/o S-I    & 79.91 & 76.14 & 73.82 & 88.17 & 36.24 & 35.53 \\
        \rowcolor{gray!10} \multicolumn{7}{c}{\textbf{Training Data}} \\
        w/o Clean  & 77.69 & 75.36 & 72.41 & 87.67 & 30.99 & 33.21 \\
        w/o Real   & 81.87 & 78.15 & 76.29 & 92.84 & 38.62 & 39.62 \\
        \midrule
        \rowcolor{green!10} Full & \textbf{82.31} & \textbf{79.20} & \textbf{77.12} & \textbf{93.10} & \textbf{41.66} & \textbf{43.14} \\
        \bottomrule
    \end{tabular}
    \vspace{-0.3cm}
\end{table}

 \subsubsection{Text Embeddings}
\label{subsubsec:text_ablation}
We train two variant models with the following configurations:
i) \textbf{w/o LTE:} We discard the learnable text embeddings [LTE] and instead use fixed templates (i.e., [cls] for clean and [cls][artifact] for artifact) as text prompts.
ii) \textbf{w/o CLS:} Object-level classification descriptions are removed from the prompts. The model uses only the templates [LTE] for clean and [LTE][artifact] for artifact without high-level semantic context.

\textbf{Analysis.}
The results are summarized in Table \ref{tab:ablation_study}. 
Replacing LTE with fixed templates (\textbf{w/o LTE}) causes a significant drop in performance (S-AUROC decreases by 5.34\%). This validates that fixed templates lack the flexibility to capture the nuances of diverse artifacts, whereas our LTE strategy allows the model to adapt the text space distribution effectively during training.
Additionally, removing the object-level descriptions (\textbf{w/o CLS}) leads to a severe degradation in classification capability (C-AUROC drops by 2.12\%). This confirms that incorporating object‑specific context facilitates accurate artifact localization, as it helps the model understand the relationship between the object's expected appearance and the anomalous artifact regions.

\subsubsection{Vision Encoder}
\label{subsubsec:vit_ablation}
We train two variant models to analyze the vision encoder architecture:
i) \textbf{w/o AD:} The adapter modules are removed. The output features of each layer in the vision encoder are passed directly to the subsequent layer without feature enhancement.
ii) \textbf{w/o MG:} The projector modules are removed, so the model relies exclusively on the final-layer global features from the vision encoder for prediction, without multi-granularity features.

\textbf{Analysis.}
As shown in Table \ref{tab:ablation_study}, removing the adapters (\textbf{w/o AD}) leads to a noticeable decline in both classification and segmentation accuracy. 
This stems from the inherent limitation of frozen CLIP features, which are optimized for high-level semantic understanding rather than fine-grained artifact discrimination. By inserting lightweight adapters, we enable task-specific feature modulation without compromising the prior knowledge encoded in CLIP. 
In addition, removing multi-granularity features (\textbf{w/o MG}) significantly impairs segmentation precision (S-F1 drops by 8.42\%). This underscores that shallow layer features are indispensable for recognizing fine-grained spatial details. 

\begin{table}[htbp]
\centering
\caption{Comparison of different methods and our variant models on three artifact types.}
\renewcommand{\arraystretch}{1.15}
\begin{tabular}{l c c c c c c}
\toprule
\multirow{2}{*}{\textbf{Method}} & \multicolumn{2}{c}{\textbf{Ghosting}} & \multicolumn{2}{c}{\textbf{Lens Flare}} & \multicolumn{2}{c}{\textbf{Moiré}} \\
\cmidrule(lr){2-3} \cmidrule(lr){4-5} \cmidrule(lr){6-7}
& \textbf{C-AP} & \textbf{S-F1} & \textbf{C-AP} & \textbf{S-F1} & \textbf{C-AP} & \textbf{S-F1} \\
\midrule
\rowcolor{gray!10}
\multicolumn{7}{c}{\textbf{Anomaly Detection Methods}} \\
MambaAD \cite{mambaad} & 48.74 & 18.37 & 56.13 & 17.78 & 73.39 & 34.26 \\
ViTAD \cite{ViTAD}   & 47.19 & 21.12 & 64.21 & 20.06 & 79.11 & 34.54 \\
UniAD  \cite{uniAD}  & 51.28 & 23.13 & 57.97 & 20.44 & 86.83 & 45.83 \\
Dinomaly \cite{dinomaly} & 54.44 & 25.94 & 59.34 & 22.97 & 82.68 & 47.21 \\
\addlinespace[0.5em]
\rowcolor{gray!10}
\multicolumn{7}{c}{\textbf{Manipulation Detection Methods}} \\
PSCCNet \cite{psccnet}  & 52.82 & 20.74 & 57.46 & 20.11 & 83.73 & 38.41 \\
MVSS-Net \cite{mvss} & 50.22 & 21.34 & 57.98 & 20.80 & 86.02 & 35.14 \\
LGrad  \cite{lgrad}  & 55.87 & 22.78 & 60.87 & 20.63 & 85.55 & 40.26 \\
FreqNet \cite{freqnet} & 57.05 & 25.76 & 60.96 & 21.44 & 85.06 & 46.19 \\
\addlinespace[0.5em]
\rowcolor{gray!10}
\multicolumn{7}{c}{\textbf{Ablation Variants}} \\
w/o LTE    & 58.46 & 25.79 & 63.70 & 23.49 & 88.95 & 46.18 \\
w/o CLS    & 67.26 & 31.82 & 71.56 & 29.88 & 95.36 & 51.85 \\
w/o AD     & 58.94 & 23.88 & 59.21 & 22.75 & 87.17 & 45.74 \\
w/o MG     & 62.69 & 25.97 & 66.74 & 24.49 & 93.26 & 53.70 \\
\midrule
\textbf{IPAD-CLIP} & \cellcolor{green!10}\textbf{68.74} & \cellcolor{green!10}\textbf{35.98} & \cellcolor{green!10}\textbf{72.22} & \cellcolor{green!10}\textbf{34.64} & \cellcolor{green!10}\textbf{96.64} & \cellcolor{green!10}\textbf{58.80} \\
\bottomrule
\end{tabular}
\label{tab:per-class}
\vspace{-0.3cm}
\end{table}

\subsubsection{Training Strategies}
To validate the effectiveness of our proposed three-stage training strategy, we train two variant models with the following configurations:
i) \textbf{w/o S-II-III}: We eliminate Stage II and Stage III, retaining only Stage I. In this setting, a classification head and a segmentation head are used for prediction, without any text guidance or multi-modal alignment.
ii) \textbf{w/o S-I}: We remove the initial adaptation of the vision encoder (Stage I) and implement only Stage II and Stage III. 

\textbf{Analysis.}
As presented in Table \ref{tab:ablation_study}, removing the multi-modal alignment mechanism (\textbf{w/o S-II-III}) results in substantial performance degradation across all metrics. This variant frequently misidentifies artifacts, leading to a high rate of false positives. Although the impact of removing Stage I (\textbf{w/o S-I}) is less severe, we observe an obvious decline in segmentation performance. This indicates that the visual adaptation performed in Stage I is crucial for refining feature representations, thereby promoting the understanding and alignment of fine-grained artifact details.
 
\subsubsection{Training Data}
\label{subsubsec:data_ablation}

We train two variant models to assess the importance of data composition:
i) \textbf{w/o Clean:} Clean images are removed from the training set, leaving only artifact images.
ii) \textbf{w/o Real:} Real-world captured images are removed from the training set, retaining only generated artifact images and clean images.

\textbf{Analysis.}
The results in Table \ref{tab:ablation_study} reveal that removing clean images (\textbf{w/o Clean}) results in a more pronounced performance decline. This is primarily because clean images serve as a vital reference, facilitating the contrast between clean and artifact images.
Although removing real-captured images (\textbf{w/o Real}) has a relatively minor impact on the metrics, it significantly degrades the model's generalization capability. We provide a more in-depth analysis and validation of this observation in the following section.

In summary, the full integration of learned text embeddings, enhanced visual features, and comprehensive training strategies and data yields optimal performance. Removing any component consistently results in degradation, validating their essential roles. This demonstrates that our framework effectively synergizes multi-modal guidance with multi-scale visual analysis to address the complex challenges of visual artifact detection.

\subsection{Discussion}

\subsubsection{Analysis on Each class.}
Table \ref{tab:per-class} presents the performance of existing methods, our IPAD-CLIP, and variant models across three artifact types. Our IPAD-CLIP consistently achieves the highest scores across all three artifact categories.
We observe that all methods perform worst on ghosting, likely because ghosting often appears along object boundaries and blends with the foreground. For lens flare, all methods exhibit low segmentation accuracy due to its ambiguous boundaries, which complicate precise localization. In contrast, moiré patterns achieve the highest accuracy among the three categories, as they typically appear on screens that are more easily distinguishable.

Removing object-level descriptions (w/o CLS) shows minimal impact on ghosting but substantial degradation on moiré. This indicates that ghosting has a weak semantic relationship with object categories, appearing as boundary effects largely independent of objects, while moiré exhibits a strong correlation, typically occurring on specific surfaces such as screens.



\begin{table}[htbp]
\centering
\caption{Generalization evaluation from synthetic data to real-captured data. }
\label{tab:generalize}
\renewcommand{\arraystretch}{1.15}
\begin{tabular}{l ccc ccc}
\toprule
\multirow{2}{*}{\textbf{Method}} & \multicolumn{3}{c}{\textbf{Synthetic}} & \multicolumn{3}{c}{\textbf{Real-captured}} \\
\cmidrule(lr){2-4} \cmidrule(lr){5-7}
 & \textbf{C-AP} & \textbf{S-AP} & \textbf{S-F1} & \textbf{C-AP} & \textbf{S-AP} & \textbf{S-F1} \\
\midrule
\textbf{Dinomaly} \cite{dinomaly} & 79.96 & 40.24 & 41.75 & 51.29 & 21.74 & 22.31 \\
\textbf{FreqNet} \cite{freqnet} & 
\cellcolor{yellow!25} 80.01 &\cellcolor{yellow!25} 42.40 & \cellcolor{yellow!25} 42.83 & 
\cellcolor{yellow!25} 58.26 & \cellcolor{yellow!25} 22.03 &\cellcolor{yellow!25} 24.19 \\
\textbf{IPAD-CLIP} & \cellcolor{green!10}\textbf{86.65} & \cellcolor{green!10}\textbf{49.03} & \cellcolor{green!10}\textbf{48.74} & \cellcolor{green!10}\textbf{67.64} & \cellcolor{green!10}\textbf{26.20} & \cellcolor{green!10}\textbf{29.48} \\
\bottomrule
\end{tabular}
\vspace{-0.3cm}
\end{table}

\subsubsection{Generalization Evaluation.}
To evaluate generalization from synthetic to real artifacts, we remove all real-captured images from the training set and reserve them for testing, resulting in two test subsets: synthetic and real-captured artifacts. We retrain Dinomaly \cite{dinomaly}, FreqNet \cite{freqnet}, and IPAD-CLIP on synthetic data and evaluate each on both subsets.
Table \ref{tab:generalize} shows that IPAD-CLIP achieves the highest scores on both subsets. This advantage is particularly pronounced on real-captured data, demonstrating its superior generalization capability.


\section{Conclusion}
In this work, we introduced the task of Image Perceptual Artifact Detection and constructed the first multi-class dataset to facilitate research in this direction. We further proposed IPAD-CLIP, a CLIP-based framework leveraging multi-modal guidance for artifact detection. Learnable text prompts encode artifact-aware semantics to align with multi-granularity image features, enhancing fine-grained artifact perception. A three-stage training strategy progressively aligns vision and text representations, shifting attention from object semantics to fine-grained artifacts. Extensive experiments demonstrate that IPAD-CLIP consistently outperforms existing detection methods, and generalization experiments confirm robust transfer from synthetic to real-captured data.

\bibliographystyle{unsrt}  
\bibliography{references}  






\end{document}